\documentclass[letterpaper, 10 pt, conference]{ieeeconf}  
\IEEEoverridecommandlockouts                              
\usepackage{graphics} 
\usepackage{epsfig} 
\usepackage{times} 
\usepackage{amsmath} 
\usepackage{amssymb}  
\usepackage{algorithm}
\usepackage[noend]{algpseudocode}
\usepackage{xcolor}
\usepackage{balance}
\usepackage{caption}
\usepackage{subcaption} 
\usepackage{cite}

\title{\LARGE \bf
Adversarial Feature Training\\ for Generalizable Robotic Visuomotor Control
}

\author{Xi Chen$^{1}$, Ali Ghadirzadeh$^{1,2}$, M{\aa}rten Bj{\"o}rkman$^{1}$ and Patric Jensfelt$^{1}$
\thanks{
\newline
$^{1}$RPL, KTH Royal Institute of Technology, Sweden \newline
$^{2}$Intelligent Robotics research group, Aalto University, Finland \newline
{\tt\small [xi8][algh][celle][patric]@kth.se}}
}

\begin{document}

\maketitle
\thispagestyle{empty}
\pagestyle{empty}

\begin{abstract}

Deep reinforcement learning (RL) has enabled training action-selection policies, end-to-end, by learning a function which maps image pixels to action outputs. 
However, it's application to visuomotor robotic policy training has been limited because of the challenge of large-scale data collection when working with physical hardware. 
A suitable visuomotor policy should perform well not just for the task-setup it has been trained for, but also for all varieties of the task, including novel objects at different viewpoints  surrounded by task-irrelevant objects. 
However, it is impractical for a robotic setup to sufficiently collect interactive samples in a RL framework to generalize well to novel aspects of a task. 

In this work, we demonstrate that by using adversarial training for domain transfer, it is possible to train visuomotor policies based on RL frameworks, and then transfer the acquired policy to other novel task domains. 
We propose to leverage the deep RL capabilities to learn complex visuomotor skills for uncomplicated task setups, and then exploit transfer learning to generalize to new task domains provided only still images of the task in the target domain. 
We evaluate our method on two real robotic tasks, picking and pouring, and compare it to a number of prior works, demonstrating its superiority. 
\end{abstract}

\section{INTRODUCTION}
Recently, end-to-end training of the perception system  jointly with the motor control part of a deep action-selection policy  has shown remarkable success to solve a wide variety of robotic problems.
Deep reinforcement learning (RL) has become a popular framework to train visuomotor action-selection policies that directly map raw  image pixels to motor actions, eliminating the need for large-scale image labeling.

However, lack of generality is still a common problem when training a visuomotor action-selection policy. 
As an example, a policy trained to pour liquid into a mug would fail if the mug is replaced by a glass that has not been seen during the training. Similar failures might occur when changing the scene, e.g., when adding novel visual clutter to the background. 
Interactive RL training is generally very costly and requires a considerable number of training samples to converge. Therefore, it is not practical to train a single robotic policy for different variations of task objects and contexts \cite{chen2018deep}. 
An appealing solution to this problem, which is suggested by Singh et al.,  \cite{singh2017gplac}, and which we pursue in the current work, is to first train a policy in a simple task setup and then to generalize to other variations of the task using still images of task objects in different domains.

Prior work has considered mainly two different approaches to address the lack of generality when jointly training the perception and control layers of a deep network in an end-to-end fashion.
The first approach is to train the perception layers of the network separately by synthetic visual data in simulation \cite{hamalainen2019affordance, zhang2017sim, tzeng2015towards, tobin2017domain,james2017transferring}.
The discrepancies between the real-world data distribution and the simulated one is compensated by generating a wide range of data in simulation by
 randomizing task-irrelevant elements of the environment, a technique known as domain-randomization.  
However, rendering a diverse set of synthetic images is often too laborious.

The second approach is to introduce a number of auxiliary loss functions to pre-train the perception layers 
followed by end-to-end training of the entire network (see \cite{ghadirzadeh2018sensorimotor} for an overview).
Auxiliary loss functions are typically used to identify objects of interest \cite{singh2017gplac}, to reconstruct them in an autoencoder architecture \cite{finn2016deep, ghadirzadeh2017deep}, to localize them \cite{devin2018deep}, or to predict the next visual state of the objects based on the current one and the applied action \cite{jaderberg2016reinforcement}. 
However, compared to the end-to-end training approach, there is no direct way to extract visual features that are relevant to the task. As an example, features which are good at reconstructing the original images of task objects in an autoencoder architecture are not necessarily suitable for object manipulation. 

\begin{figure}[t]
\centering
\includegraphics[width=0.49\textwidth]{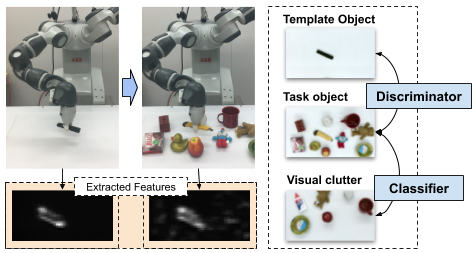}
\caption{A policy is first trained in the simplest possible setup based on a template object without clutter. The trained policy is then transferred to work with other task objects in a more realistic setup, i.e., in the presence of visual clutter.
}
\label{fig:first_pic}
\vspace{-6mm}
\end{figure}

In this work, we introduce an end-to-end training method based on adversarial training to extract visual features which generalize well to other instances of task objects, assuming that weakly labeled still images of such objects are provided.
Here, weak labels mean binary values specifying whether a given image contains a task object or not. 
The main idea, shown in Fig.~\ref{fig:first_pic}, is to train the perception system to generate similar visual features in different contexts where the same sequence of motor actions successfully completes the task.
In our experiments, we show that we can train a policy for only a template object, an object with the simplest possible geometry and texture, without adding visual clutter, and then selectively transfer the learned policy to other task objects. 
The advantages of our method are threefold: 
(1) improving the efficiency by training the most data demanding phase, i.e., the policy optimization, in an uncomplicated task setup, 
(2) circumventing the tedious and time-consuming labeling process, since unlike most prior work, our method does not need bounding box labels to determine task objects, and (3) gaining full control of what to include into the set of task objects by weakly labeling images that contain the  desired objects. 
Our empirical analysis demonstrates that our method outperforms prior work with a good margin, when 
performing the task for novel objects, objects for which only cluttered still images are provided.
We demonstrate the appropriateness of our method by training and evaluating feed-forward action-selection policies for two real robotic tasks, picking and pouring using an ABB YuMi robot. 

The organization of this paper is  as  follows:  In  the  next section,  we    review related work. Sec.~ \ref{sec:method} introduces our method. Experimental  results  are  provided  in  Sec.~\ref{sec:experiments}. Finally, Sec.~\ref{sec:conclusion} will conclude our discussion and suggest future work.
\section{RELATED WORK}
\label{sec:relatedwork}
In this section, we review prior work to extract a compact feature representation of visual inputs for learning visuomotor control policies. Also, since our method is categorized as transfer learning, we briefly describe related studies to unsupervised domain adaptation.

\subsection{Visual representation learning for visuomotor control}
Levine et al., \cite{levine2016end} proposed end-to-end training of perception and control networks for a number of robotic visuomotor tasks. However, trained policies easily fail in new task setups because of the limited number of training samples which can be realistically collected from a real robotic setup. 
Pre-training the perception model using autoencoders \cite{finn2016deep,ghadirzadeh2017deep, hamalainen2019affordance} has gained in popularity since such training does not require real robot samples. 
However, these methods may not be able to extract task-specific features as they are not trained explicitly for a given task. 

Training with auxiliary tasks which share the state representations with the primary task is another commonly used approach to acquire task-specific features \cite{jaderberg2016reinforcement,hariharan2014simultaneous,zhang2014facial,gupta2017learning}. 
In robotic manipulation applications, object detection and localization can be considered as two important auxiliary tasks to help the extraction of suitable visual features. 
Devin et al., \cite{devin2018deep} proposed a two-stage approach that first trains a task-independent attention module to attend to objects in the scene. Then, they train a task-specific network to find out the task objects provided a number of human demonstrations. Jang et al., \cite{jang2018grasp2vec} proposed a method to learn a representation to localize objects by comparing the change in the visual features when adding or removing an object. Singh et al.,  \cite{singh2017gplac}, proposed to include a binary classification task which, similar to our work, determines if the input image contains the task object or not.
We extend the work in \cite{singh2017gplac} by introducing an adversarial loss beside the classification loss. We empirically demonstrate that introducing this loss improves the performance considerably. 


\subsection{Unsupervised domain adaptation}
Unsupervised domain adaptation focuses on scenarios where labeled data are not available from the target domain.
Earlier studies on domain adaptation try to learn domain-invariant features by minimizing differences between the domain distributions given a distance measure. 
The maximum mean discrepancy (MMD) loss is used by
\cite{tzeng2014deep, rozantsev2018beyond, long2015learning} to align source and target features. \cite{sun2016return, sun2016deep} applied correlation alignment (CORAL) that tries to align the second-order statistics of the two domains. \cite{kang2019contrastive} proposed the contrastive domain discrepancy (CDD) loss which minimizes the intra-class domain discrepancy and maximizes the inter-class domain discrepancy. 
Other methods choose to learn domain-agnostic representations by using adversarial training 
\cite{ganin2014unsupervised, ganin2016domain}.
Tzeng et al., \cite{tzeng2017adversarial} proposed a general framework for adversarial deep domain adaptation (ADDA) which uses GAN \cite{goodfellow2014generative} loss to align the two domains.
Most of these methods are applied to computer vision classification tasks with  relatively small domain shifts. 
Our method combines the adversarial domain adaptation in \cite{tzeng2017adversarial} with the attention mechanism introduced in \cite{singh2017gplac} to attend to the task objects and align the feature representations of the target domain and the source domain. 
\section{Method}
\label{sec:method}

In this section, we introduce our method based on adversarial training to extract a set of visual features as the output of a perception model which generalizes well to other task related objects. 

\begin{figure}[h]
\centering
\includegraphics[width=0.45\textwidth]{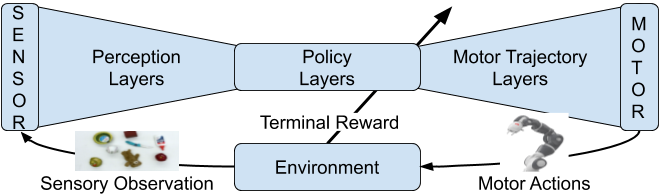}
\caption{Our deep policy network maps an input sensory data to a sequence of actions through three sub-networks: the perception network, the low-dimensional policy network and the motor trajectory network.}
\label{fig:tra_policy}
\vspace{-3mm}
\end{figure}


\subsection{Problem Formulation}
The problem we address is how to train a deep feed-forward policy network which maps an input image $I$ to a sequence of motor actions $u_{t=0:T-1}$, where $T$ is the length of the motor trajectory, and $u_t$ is the $M$-dimensional motor action at time-step $t$. 
Similar to our earlier work \cite{ghadirzadeh2017deep}, we 
split a deep policy network into three sub-networks, which are (1) the perception, (2) the low-dimensional policy, and (3) the motor trajectory networks, as shown in the Fig.~\ref{fig:tra_policy}. 

The perception network is responsible for finding a suitable representation of the input image. 
The primarily goal of this work is to find a method to train the perception model such that consistent features are extracted for different task objects. 
This would lead to a higher success rate when handling different objects, since the output of the perception network would be similar despite differences in task objects. 

The motor trajectory network is a generative model which maps a latent variable input into a trajectory of motor actions. 
This means that the sequential decision making problem is converted into a multi-armed bandit problem without temporal complexities. 
In other words, instead of searching for optimal motor actions for every time-step, a single low-dimensional latent variable (typically less than 5 dimensions) that represents the complete motor trajectory is searched for. 
Similarly to our earlier work \cite{ghadirzadeh2017deep, ghadirzadeh2018sensorimotor, hamalainen2019affordance}, variational autoencoders (VAEs) \cite{kingma2013auto} are used to find the trajectory space, with the trajectory network pre-trained and kept fixed during policy training that follows. 



The low-dimensional policy 
is implemented as a small network which maps visual features to the latent space of the generative model. 
The parameters of this network are commonly trained using reinforcement learning, e.g., the proximal policy optimization (PPO) method \cite{schulman2017proximal}. In practice, due to the reduced dimensionalities of both features and trajectories, most modern policy optimization method can be applied.
During training, an input image is first represented by a set of features when it is passed through the perception model. The policy network maps the features to a value in the latent space. The latent value is then passed to the generative model to produce a full trajectory of the motor actions. 
A terminal reward value is received at the completion of the task, determining the success of the agent to accomplish the goal. 
The policy is then trained, end-to-end, with the perception network, to maximize the expected value of the terminal reward. 


Since the main focus of this paper is on training the perception network, we 
refer to our earlier studies (\cite{ghadirzadeh2017deep,ghadirzadeh2018sensorimotor,hamalainen2019affordance}) 
for details on the training of policies and the generative motor trajectory models.

\subsection{Training Data-sets}
\label{sec:method_training_data_sets}
To avoid the expensive process of training deep visuomotor policies based on real robot data and at the same time allow policies to generalize to novel variations of the task, 
we propose to first train the policy given interactive data collected  by playing with a template object only, and then to transfer the gained manipulation knowledge to the rest of the task objects assuming that still images of those objects are provided. 
Therefore, we need to construct two sets of training data as described in the following:

\noindent
    \textbf{Task data-set (source domain data $\mathcal{D}_S$)}:
    Using reinforcement learning we train the low-dimensional policy and the perception model jointly using visual images of the template object without background clutter. 
    Once the policy training has converged, we record samples including input visual images and output motor actions. In our case, the actions are represented by the latent values. 
    The task data-set includes the input-output data of the pre-trained low-dimensional policy and the perception network.
    
\noindent
    \textbf{Task object data-set (target domain data $\mathcal{D}_T$)}: 
    Since our goal is to train a policy which generalizes well to other task objects, we collect 
    still images of such objects surrounded by visual clutter as they appear in real task setups. The task object data-set includes these images accompanied by weak labels that specify whether an image contains a task object or not.
    However, we do not specify which object it is or where the object is located in the image.  

\subsection{Adversarial Feature Training}
\label{sec:adversarial_training}

In this section, we introduce our method for end-to-end training of the perception and policy networks. 
These networks are pre-trained jointly using the RL method as explained earlier. 
In this phase, both are further trained to perform well for novel task objects. 
The problem we are addressing can be formulated as a transfer learning problem, where the primarily goal is to transfer the knowledge gained from RL training with the template object (source domain $\mathcal{D}_S$) to the rest of task objects (target domain $\mathcal{D}_T$). 
The main idea is to extract a set of visual features as the output of the perception model which are suitable for the RL task while not being distinguishable whether the input image comes from the source domain or the target domain. 

We introduce two auxiliary networks, a discriminator and a classifier, which are not part of the final model but will only be used during training. 
The discriminator network receives the visual features generated by the perception model and decides whether the input is an image 
from the source domain or from the target domain.
Similarly, the classifier network receives the visual features as input and outputs a classification of whether there is a task object in a target domain input image or not. 

We train the perception, the policy, as well as the auxiliary networks, the discriminator and the classifier, jointly by optimizing the three loss functions introduced in the following:
\begin{equation}
    \mathcal{L} = \mathcal{L}_{task}(\mathcal{D}_S) + \mathcal{L}_{c}(\mathcal{D}_T) + 
    \mathcal{L}_{D}(\mathcal{D}_S \cup \mathcal{D}_T), 
\end{equation}
where $\mathcal{L}_{task}$ is the task loss function, $\mathcal{L}_{c}$ is the binary classification loss, and $\mathcal{L}_{D}$ is the binary discriminator loss. 
Details on the losses are provided as follows:

\noindent
\textbf{Task loss:} $\mathcal{L}_{task}$ is a mean squared error (MSE) loss applied to the source domain data to learn the mapping from the template object images to the corresponding latent motor trajectory space as provided by the task data-set. The perception and the policy networks are optimized jointly to minimize the task loss function. 

\noindent
\textbf{Classification loss:}
$\mathcal{L}_{c}$ is a binary cross-entropy classification loss, similar to \cite{singh2017gplac}, which is used to classify whether the input image contains a task object or not. 
The perception and classifier networks are trained jointly using the target domain data. 
This loss function helps the perception model to attend to task objects without requiring an extra effort to explicitly locate the objects by manually drawing bounding-boxes.  

\noindent
\textbf{Discriminator loss:}
$\mathcal{L}_{D}$ is a discriminator loss that forces the extracted features to be distributed similarly irrespective of their origin, whether the features come from the source domain data or the target domain data, i.e~whether the image contains a single template object or a target object hidden in visual clutter. 

The discriminator and perception models are trained using the adversarial training paradigm. 
Given sample images from the source and the target domains, the perception model generates visual features corresponding to the input. 
The discriminator network is then trained to tell the source and the target domain data apart by observing the corresponding feature vector. 
Next, the perception network is updated to generate new visual features such that the current discriminator cannot identify whether the features come from the source domain or from the target domain.


To sum up, the task and discriminator losses are required to extract a set of visual features suitable for accomplishing the task, not only for the template object, but for the rest of the task objects when these are surrounded by visual clutter. 
The classification loss speeds up the training by implicitly guiding the network where to attend in the input image. 

\section{EXPERIMENT}
\label{sec:experiments}

In the experiment section, we demonstrate two real-world robotic manipulation tasks and evaluate our method based on (1) if our method can handle different visual clutter in the scene, and (2) if the learned policy can generalize well to different task objects, including objects that were unseen during policy training.

\subsection{Tasks setup and data collection}
\begin{figure}[t]
\centering
\includegraphics[width=0.35\textwidth]{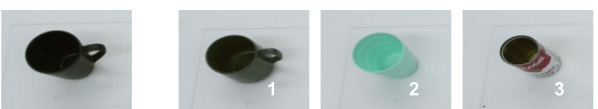}
\caption{Task objects used in the pouring task. The object on the left is the template object. The target object ID is labeled on the right bottom of the images.}
\label{fig:put_objects}
\vspace{-3mm}
\end{figure}

\begin{figure*}[t]
\centering
\includegraphics[width=0.98\textwidth]{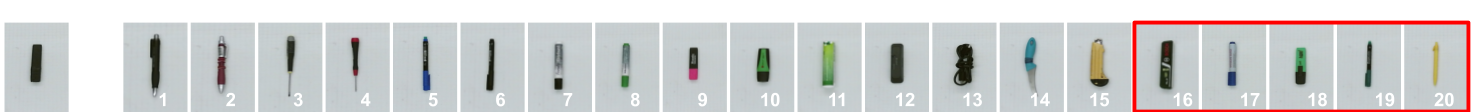}
\caption{Task objects used in the picking task. The object on the left is the template object. The object ID is labeled on the right bottom of the image. Objects from $16$-$20$ are unseen objects which are not included in the weakly labeled data-set.}
\label{fig:pick_objects}
\vspace{-3mm}
\end{figure*}

Our experiments are performed on an ABB Yumi robot. The experiments include two tasks, a pouring task and a picking task. In the pouring task, the robot needs to pour the content of a small cup into a desired cup placed on a table, and in the picking task, the robot has to grasp and lift a cuboid object from the table. The cups and the cuboid objects are placed together with other task-irrelevant objects that are regarded as visual clutter.
For each task, we first train a policy using a template object and then transfer the learned policy to other task objects, as explained in Sec.~\ref{sec:method}. 
We select $3$ cups for the pouring task and $20$ cuboid objects for the picking task. The template objects and task objects used in the experiments are shown in Fig.~\ref{fig:put_objects} and \ref{fig:pick_objects}.
As described in Section~\ref{sec:method_training_data_sets}, for each task, we first train the low-dimensional policy and perception model jointly on the template object using reinforcement learning. Then we record $500$ input RGB images of the template object in different locations and orientations, together with $500$ corresponding latent values of the low-dimensional policy as the task data-set (source domain). 

The target domain data-set contains still images of the task objects and visual clutter. To each image in this data-set, we assign a binary label. If the image contains only visual clutter, we labeled it as $1$, but if it contains a task object, we label it as $0$. Maximum one task object can appear in each image.
We collect $70$ images for each task object. Therefore, there are $3\times70=210$ images for the pouring task and $15\times70=1,050$ images for the picking task.
We collect $1,000$ images of visual clutter, which are shared between the pouring and picking tasks. 
The locations and orientations of all objects are randomly selected, as well as the number of task-irrelevant objects representing the visual clutter.

\subsection{Baseline methods}
\begin{figure}[h]
\centering
\includegraphics[width=0.45\textwidth]{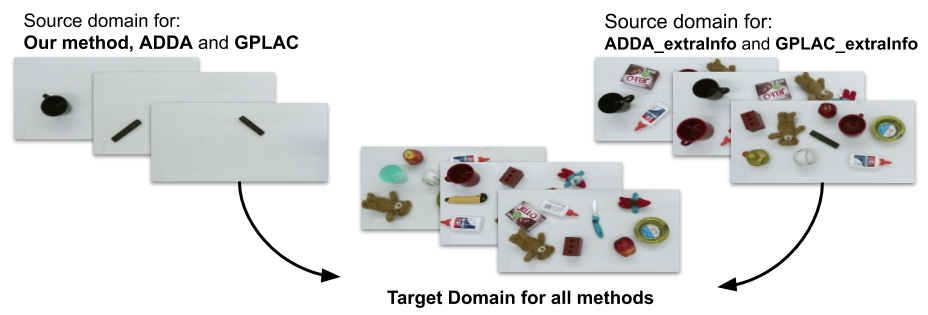}
\caption{Two task data-sets used in the experiments. We train our method, ADDA and GPLAC using the task data-set contains only the template object (top left), and we train ADDA\_extraInfo and GPLAC\_extraInfo using the task data-set contains both the template object and visual clutter (top right). All methods should adapt to the same target domain that contains task object and visual clutter (middle). 
}
\label{fig:train_info}
\end{figure}

We compare our method to two baseline methods closely related to ours: ADDA \cite{tzeng2017adversarial} and GPLAC \cite{singh2017gplac}. ADDA is an unsupervised domain adaptation method which uses a task loss and a discriminator loss to update the network. GPLAC is a method that learns generalizable robotic skills from weakly labeled data using task and classification losses, while our method uses a combination of task, discriminator and classification losses.
However, since neither baseline method is able to perform satisfactorily on the task data-set that we have, we prepare another task data-set where the template object is placed together with task-irrelevant objects. We thus reduce the complexity of the task by providing extra information of the irrelevant objects, which reduces the discrepancy between the source and target domains. 
We train two more models of ADDA and GPLAC using the task data-set with visual clutter and refer to these as ADDA\_extraInfo and GPLAC\_extraInfo. Fig.~\ref{fig:train_info} illustrates the training setup for each method.

There are some implementation differences between the baseline methods used in our work and that of the original paper. 
For ADDA in \cite{tzeng2017adversarial}, the perception network for the source and target domains are separate. In our work, we found that sharing the weights of the perception network generates better performance, especially for ADDA\_extraInfo. Thus, we train ADDA and ADDA\_extraInfo using a shared perception network for both source and target domains.    
As described in \cite{singh2017gplac}, GPLAC applies a spatial softmax layer after the convolution layer to extract the position of points of maximal activation in each channel.
In order to extract the spatial feature, we increase the number of channels from $1$ to $16$ and extract $16$ points as spatial features, which are then feed to the rest of the network.  

\begin{figure*}[h]
\centering
\includegraphics[width=0.95\textwidth]{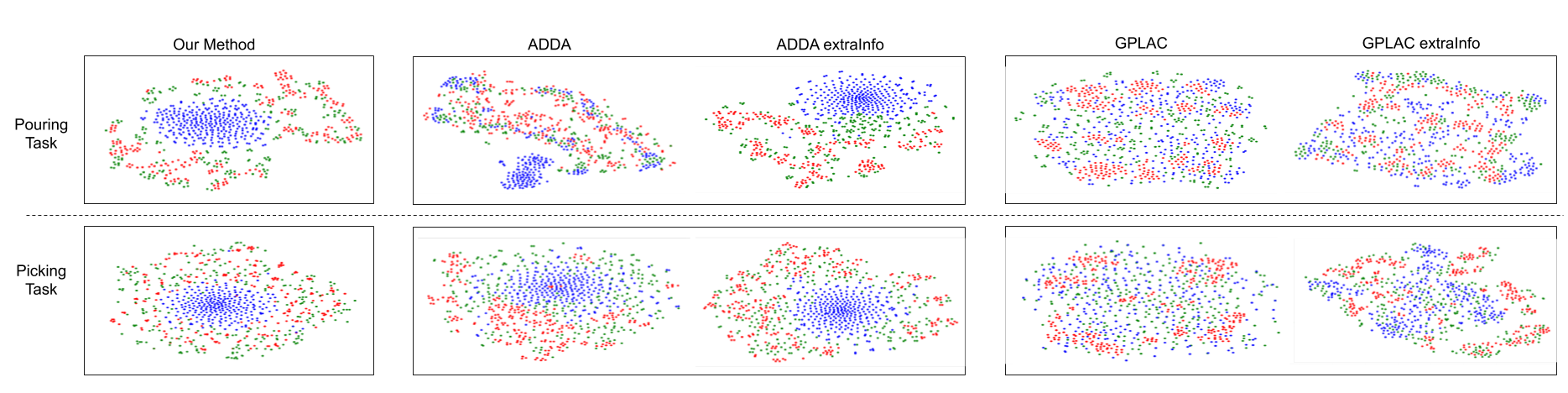}
\caption{The distribution of visual features extracted by different methods. The distribution of data used for pouring task is shown in the first row and the distribution for picking task is shown in the second row.  
The dimension of the feature is reduced using t-SNE. The features of images contain template object are plotted in red, those contain task object are plotted in green, and data of visual clutter are plotted in blue. 
}
\label{fig:tsne_all}
\vspace{-3mm}
\end{figure*}

\subsection{Experiment results}
We perform $10$ tests for each task object, which is placed in different random locations and orientations with different configurations of visual clutter,
and assign a score to represent the completion of the task.
For the pouring task, we assign $0$ when the content falls outside the desired cup and $1$ when it falls inside. For the picking task, we assign $0$ when the robot fails to grasp or lift the task object, $0.5$ when the robot successfully lifts it but with an unstable grasping pose (e.g. grasp the edge of the object), and $1$ when the robot successfully lifts it with a grasping pose close to the center of object. We measure the average score as the success rate and list the experimental result of the pouring task in Table~\ref{table:put_result} and the picking task in Table~\ref{table:pick_result}. Additionally, in Table~\ref{table:pick_result_unseen}, we report the picking performance on $5$ objects that were not seen during training. 
The results of ADDA and GPLAC on the pouring task and the result of GPLAC on the picking task are not reported as we failed to learn a working policy.

\subsubsection{Learning to ignore visual clutter}
\begin{table}[]
\centering
\caption{Success rate of pouring task on target objects}
\label{table:put_result}
\begin{tabular}{cccccc}
\hline
\textbf{object ID}                                                     & \textbf{Ours} & \textbf{\begin{tabular}[c]{@{}c@{}}ADDA\\ extraInfo\end{tabular}} & \textbf{\begin{tabular}[c]{@{}c@{}}GPLAC\\ extraInfo\end{tabular}} \\ \hline
1              & \textbf{1.0}             & 0.9                            & \textbf{1.0}    \\
2              & \textbf{1.0}             & 0.0                            & 0.0                                                                  \\
3              & \textbf{1.0}             & 0.7                            & 0.1                                                                \\ \hline
\textbf{\begin{tabular}[c]{@{}c@{}}Average\\ Success rate\end{tabular}} 
              & \textbf{1.00}             & 0.53                           & 0.37                                                               \\ \hline
\end{tabular}
\end{table}
\begin{figure}[h]
\centering
\includegraphics[width=0.49\textwidth]{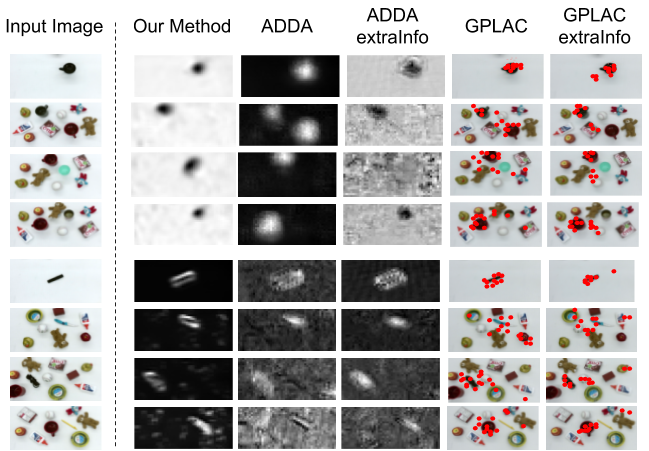}
\caption{Examples of visual feature extracted by different methods. The image in the first column is the input RGB data and the rest of the images are the extracted feature from the input image. For GPLAC and GPLAC\_extraInfo, we plot the extracted points on the input images.   
}
\label{fig:example}
\end{figure}

Our method achieves $100\%$ success rate on the pouring task while ADDA and GPLAC fail to complete the task.
In the pouring task, the model should recognize the task object while ignoring the visual clutter including objects that look very similar to the template object. This task is difficult for ADDA as the model easily misinterprets a cup in the visual clutter as the target cup. In Fig.~\ref{fig:example}, we can see that ADDA extracts stronger features on the visual clutter cup than the task cup, which leads the policy to attend to the wrong cup. 
The task is also challenging for GPLAC. Even with the classification loss on the weakly labeled data-set, which helps the network to attend to features on the task object, the features are not strong enough to eliminate the influence of visual clutter in the scene. The perception network is not able to generate features consistent between the source and target domains, resulting in the policy not showing a desired behavior on the task object. In Fig.~\ref{fig:example}, we see that the spatial features extracted by GPLAC are spread over the entire image.

When we add task-irrelevant objects to the source domain for training, both ADDA\_extraInfo and GPLAC\_extraInfo result in improved performance, especially for cup 1 in the pouring task (in Fig.~\ref{fig:put_objects}) which is similar to the template cup. However, they still fail to generalize to objects with a different color than the template object, such as cup 2. From Fig.~\ref{fig:example} we can see that ADDA\_extraInfo generates stronger features on cup 1 and cup 3, but could not detect cup 2. GPLAC\_extraInfo also detects cup 1 successfully but fails on cup 2 and cup 3. 

For better visualization of the distribution of each domain, we plot the extracted features in Fig.~\ref{fig:tsne_all}. The dimension of features is reduced to $2$ using t-SNE \cite{maaten2008visualizing}. Cases containing template object are plotted in red, those containing task objects are plotted in green, data containing pure task-irrelevant objects are plotted in blue. In the ideal case, the red points should mix with the green points, meaning that the features extracted from the source domain (red points) are well-aligned with the features extracted from the target domain (green points). The blue points should be well separated from both red points and green points, meaning that we are not using features from the task-irrelevant to learn the policy. It is clear from Fig.~\ref{fig:tsne_all} that the blue points and the green points of ADDA are mixed, which indicates ADDA extracts features from the wrong objects. After adding extra information to the source domain, the alignment/separation are improved. The result of GPLAC is not as clear as other methods, as it uses $16$ spatial point feature instead of image features. However, we can still observe improvements in local areas.  

\subsubsection{Generalizing to similar targets}
\begin{table}[]
\centering
\caption{Success rate of picking task on target objects}
\label{table:pick_result}
\begin{tabular}{cccccc}
\hline
\textbf{object ID}                                                     & \textbf{Ours} & \textbf{ADDA} & \textbf{\begin{tabular}[c]{@{}c@{}}ADDA\\ extraInfo\end{tabular}} & \textbf{\begin{tabular}[c]{@{}c@{}}GPLAC\\ extraInfo\end{tabular}} \\ \hline
1                                                                      & \textbf{1.00} & 0.70          & 0.85                                                              & 0.85                                                               \\
2                                                                      & \textbf{0.90} & 0.25          & 0.60                                                              & 0.05                                                               \\
3                                                                      & \textbf{0.95} & 0.60          & 0.50                                                              & 0.50                                                               \\
4                                                                      & \textbf{0.90} & 0.40          & 0.55                                                              & 0.45                                                               \\
5                                                                      & \textbf{0.90} & 0.30          & 0.80                                                              & 0.10                                                               \\
6                                                                      & \textbf{1.00} & 0.25          & 0.70                                                              & 0.95                                                               \\
7                                                                      & \textbf{0.80} & 0.50          & 0.80                                                              & 0.40                                                               \\
8                                                                      & \textbf{0.85} & 0.60          & 0.65                                                              & 0.30                                                               \\
9                                                                      & 0.80          & 0.25          & 0.40                                                              & \textbf{1.00}                                                      \\
10                                                                     & \textbf{0.90} & 0.45          & 0.80                                                              & 0.60                                                               \\
11                                                                     & 0.80          & 0.15          & \textbf{0.85}                                                     & 0.35                                                               \\
12                                                                     & 0.90          & 0.00          & 0.90                                                              & \textbf{0.95}                                                      \\
13                                                                     & 0.75          & 0.00          & 0.75                                                              & \textbf{0.80}                                                      \\
14                                                                     & \textbf{0.85} & \textbf{0.85}          & 0.60                                                              & 0.00                                                               \\
15                                                                     & \textbf{0.65} & 0.00          & 0.35                                                              & 0.00                                                               \\ \hline
\textbf{\begin{tabular}[c]{@{}c@{}}Average\\ Success rate\end{tabular}} & \textbf{0.86} & 0.35          & 0.67                                                             & 0.49                                                               \\ \hline
\end{tabular}
\end{table}

\begin{table}[]
\centering
\caption{Success rate of picking task on unseen objects}
\label{table:pick_result_unseen}
\begin{tabular}{cccccc}
\hline
\textbf{object ID}                                                     
& \textbf{Ours} & \textbf{ADDA} & \textbf{\begin{tabular}[c]{@{}c@{}}ADDA\\ extraInfo\end{tabular}} & \textbf{\begin{tabular}[c]{@{}c@{}}GPLAC\\ extraInfo\end{tabular}} \\ \hline
16   & 0.75                    & 0.10           & \textbf{1.00}               & \textbf{1.00}                                                                  \\
17   & \textbf{0.70}           & 0.20           & 0.55                        & 0.00                                                                  \\
18   & \textbf{0.85}           & 0.55           & 0.45                        & 0.65                                                                \\
19   & \textbf{0.90}           & 0.65           & 0.80                        & 0.60                                                                \\
20   & \textbf{0.90}           & 0.00           & 0.00                      & 0.00                                                                  \\ \hline
\textbf{\begin{tabular}[c]{@{}c@{}}Average\\ Success rate\end{tabular}} & \textbf{0.82}          & 0.30          & 0.56                                                              & 0.45                                                               \\ \hline
\end{tabular}
\end{table}
In the picking task, the model should apply the picking motion learned from the template object to a variety of other objects that are different in terms of size, color and texture (Fig.~\ref{fig:pick_objects}). 
In this experiment, ADDA performs better than for the previous task, as the task objects are not of the same shape as the task-irrelevant objects. 
However, due to the same reason discussed in the previous task, GPLAC still suffers from confusion with the task-irrelevant objects. 
Similar to the pouring task, adding extra information to the source domain considerably improves the performance. However, compared to our method, the ability of generalize is still limited for the baseline methods.
As reported in Table~\ref{table:pick_result} and Table~\ref{table:pick_result_unseen}, we achieve a success rate of over $80\%$ on most task objects while the baseline methods, even with extra training information, only perform better on a few objects.
The visualization of the feature distribution is shown in Fig.~\ref{fig:tsne_all}, where our method demonstrates a better alignment/separation performance than the baseline methods.




\section{CONCLUSIONS}
\label{sec:conclusion}
In this paper, we introduced a method to train visuomotor robotic control policies which generalize well to unseen task domains. 
We formulated the problem as a transfer learning problem in which the labels (correct motor actions) are missing for the target domain data, i.e., images of the task objects.  
We demonstrated that we can train a policy for an uncomplicated task setup using a template object, and then generalize to novel task domains using the adversarial learning approach and the auxiliary classification task introduced in this work. 
In this way, there is no need for RL interactive training by collecting a huge number of real robot data. Furthermore, tedious image labeling is also avoided since, similar to the prior work, we only need weakly labeled images. 
We evaluated our method on two real robotic tasks, pouring and picking, and compared it to two baseline methods.
The experimental results demonstrated that our method achieves considerably better performance to successfully accomplish the tasks with novel task objects.

As a part of our future work, we are interested in applying the method to more elaborate tasks which require better scene understanding to manipulate objects, e.g., pushing  objects while avoiding obstacles.
Also, our method can be applied to sim-to-real transfer learning to learn a task completely in simulation with a number of template objects and then to transfer the policy to manipulate real objects.

\section{ACKNOWLEDGMENTS}
 This work is supported by the European Union’s Horizon 2020 research and innovation program, the socSMCs project (H2020-FETPROACT-2014), and also by the Academy of Finland through the DEEPEN project.

\bibliographystyle{IEEEtran}
\balance
\bibliography{root}

\end{document}